%% file: sec0_main.tex
\documentclass[sigconf]{acmart}
\AtBeginDocument{%
  }

\setcopyright{acmlicensed}

\copyrightyear{2026}
\acmYear{2026}
\setcopyright{cc}
\setcctype{by}
\acmConference[MM '26]{Proceedings of the 34th ACM International Conference on Multimedia}{November 10--14, 2026}{Rio de Janeiro, Brazil}
\acmBooktitle{Proceedings of the 34th ACM International Conference on Multimedia (MM '26), November 10--14, 2026, Rio de Janeiro, Brazil}
\acmDOI{10.1145/3767308.3835986}
\acmISBN{979-8-4007-2213-4/2026/11}

\usepackage{amsmath}
\usepackage{mathtools}
\usepackage{amsthm}

\usepackage[capitalize,noabbrev]{cleveref}
\theoremstyle{plain}

\theoremstyle{definition}

\theoremstyle{remark}

\usepackage[textsize=tiny]{todonotes}

\usepackage[utf8]{inputenc} %
\usepackage[T1]{fontenc}    %
\usepackage{hyperref}       %
\usepackage{url}            %
\usepackage{booktabs}       %
\usepackage{amsfonts}       %
\usepackage{nicefrac}       %
\usepackage{microtype}      %
\usepackage{xcolor}         %

\usepackage{natbib}  %
\usepackage{caption} %
\usepackage{algorithm}
\usepackage{listings}
\usepackage{multirow}
\usepackage[outdir=./]{epstopdf}
\usepackage{enumitem}
\usepackage{subcaption}
\usepackage{subfloat}
\usepackage{newfloat}
\usepackage{graphicx}
\usepackage{svg} %
\usepackage[normalem]{ulem}
\usepackage{framed}
\usepackage{mdframed}
\usepackage{lipsum}
\usepackage{float}
\usepackage{xurl}
\usepackage{makecell}
\usepackage{array}
\usepackage{multirow}
\usepackage{wrapfig}
\usepackage{makecell}
\usepackage{enumitem}
\usepackage{pifont}
\newcommand{\cmark}{\ding{51}} %
\newcommand{\xmark}{\ding{55}} %
\usepackage{algorithmic}
\usepackage{setspace}
\usepackage{float}
\usepackage{tcolorbox}
\usepackage{colortbl}
\usepackage[table]{xcolor} %
\definecolor{lightgray}{rgb}{0.94, 0.94, 0.94}
\definecolor{lightblue}{rgb}{0.85, 0.92, 1.0}
\definecolor{lightgreen}{rgb}{0.8, 1.0, 0.8}
\definecolor{lightred}{rgb}{1.0, 0.85, 0.85}
\definecolor{lightyellow}{rgb}{1.0, 0.93, 0.80}
\definecolor{lightpurple}{rgb}{0.92, 0.88, 0.98}

\begin{document}

\title{CityRiSE: Reasoning Urban Socio-Economic Status in Large Vision-Language Models via Reinforcement Learning}

\author{Tianhui Liu}
\affiliation{%
 \institution{Information Hub, \\The Hong Kong University of Science and Technology (Guangzhou)}
 \country{Guangzhou, China}
 \institution{\\Tsinghua University}
 \country{Beijing, China}
}
\email{tianhui.liu@connect.hkust-gz.edu.cn}

\author{Hetian Pang}
\affiliation{%
 \institution{Department of Electronic Engineering, BNRist,\\Tsinghua University}
 \country{Beijing, China}
}
\email{pht23@mails.tsinghua.edu.cn}

\author{Xin Zhang}
\affiliation{%
  \institution{Department of Electronic Engineering, BNRist,\\Tsinghua University}
 \country{Beijing, China}
}
\email{zhangxin4087@163.com}

\author{Jie Feng$\dagger$}
\affiliation{%
 \institution{Zhongguancun Academy}
 \country{Beijing, China} 
 \institution{\\Tsinghua University}
 \country{Beijing, China}
}
\email{fengjie@bza.edu.cn}

\author{Pan Hui$\dagger$}
\affiliation{%
  \institution{Information Hub, \\The Hong Kong University of Science and Technology (Guangzhou)}
 \country{Guangzhou, China}
 }
\email{panhui@ust.hk}

\author{Yong Li$\dagger$}
\affiliation{%
  \institution{Department of Electronic Engineering, BNRist,\\Tsinghua University}
 \country{Beijing, China}
 }
\email{liyong07@tsinghua.edu.cn}

\renewcommand{\shortauthors}{Tianhui Liu et al.}

\begin{abstract}
  Urban socio-economic sensing plays a vital role in advancing global sustainable development goals. With the advent of Large Vision-Language Models (LVLMs), new opportunities have emerged to address this challenge by framing it as a multi-modal perception and reasoning task. However, recent studies show that LVLMs still struggle to make accurate and interpretable socio-economic predictions from visual data. To overcome these limitations and fully exploit the potential of LVLMs, we propose CityRiSE, a novel framework for Reasoning urban Socio-Economic status in LVLMs via reinforcement learning (RL). With carefully curated multi-modal dataset and verifiable reward design, our approach guides the LVLM to focus on semantically meaningful visual cues, enabling structured and goal-oriented reasoning for generalist socio-economic status prediction. Experiments demonstrate that CityRiSE, equipped with emergent reasoning,  significantly outperforms existing baselines, improving both prediction accuracy and generalization across diverse urban contexts, especially on unseen cities and unseen indicators. This work highlights the promise of combining RL and LVLMs for interpretable and generalist urban socio-economic sensing.
\end{abstract}


\maketitle
\begingroup
\renewcommand{\thefootnote}{\fnsymbol{footnote}}
\footnotetext[2]{Corresponding authors.}
\endgroup

\input{sec1_intro}

\input{sec2_methods}
\input{sec3_experiment}

\input{sec4_related_work}
\vspace{-5pt}
\section{Conclusion}\label{sec:conclusion}
In this paper, we propose CityRiSE, the first framework to achieve generalist urban socio-economic status prediction across diverse cities and indicators by leveraging the emergent reasoning capabilities of LVLMs through reinforcement learning. 
CityRiSE first achieve cross-city and cross-indicator generalization by training the model on normalized prediction targets, a process that encourages learning transferable reasoning patterns connecting visual input to high-level socio-economic concepts.
By combining GRPO with two auxiliary datasets that enhance transferable perceptual and logical reasoning skills, CityRiSE achieves strong performance using only 5k training samples, demonstrating high data efficiency.
Finally, we design a verifiable multi-component reward mechanism that induces coherent and goal-directed reasoning, thereby enhancing both interpretability and prediction quality.
Extensive experimental results demonstrate the superiority of CityRiSE over SOTA baselines for 10 urban socio-economic prediction tasks.

\begin{acks}
This work was supported in part by the National Key Research and Development Program of China under Grant 2024YFC3307602, the Guangdong Provincial Talent Program under Grant No. 2023JC10X009, and the National Key Research and Development Program of China under Grant 2024YFC3307603. It was also sponsored by the Tsinghua-Toyota Joint Research Institute Inter-disciplinary Program (Toyota-Future City Simulator). Additional support was provided the Zhongguancun Academy under Grant No. XTS0074.
\end{acks}

\bibliographystyle{ACM-Reference-Format}
\bibliography{reference}

\end{document}

%% file: sec1_intro.tex
\section{Introduction} \label{sec:introduction}
Socio-economic indicators, such as GDP, population density, education level, and healthcare access, are critical for assessing the development, equity, and sustainability of urban regions. These metrics play a foundational role in guiding urban planning, policy-making, and international initiatives such as the United Nations Sustainable Development Goals (UN SDGs), which emphasize the importance of data-driven insights for fostering sustainable and inclusive growth.
In recent years, there has been growing interest in predicting these indicators using externally observable data sources, in order to provide timely and spatially detailed insights that go beyond the limitations of traditional census-based methods. 
However, this task is inherently challenging. Socio-economic indicators are abstract, high-level targets that must be inferred from complex, heterogeneous, and often noisy data sources, with satellite imagery and street view images being among the most widely used modalities. Extracting informative features from such multi-modal inputs and effectively mapping them to socio-economic outcomes remains a non-trivial problem. 

\begin{table}[t]
\centering
\renewcommand\cellalign{lc}
\renewcommand\theadalign{lc}
\setlength{\tabcolsep}{2pt} %
\caption{Comparison of Methods on City Transferability, Indicator Generalization, Data Scale, and Reasoning Ability.}
\vspace{-10pt}
\label{tab:method_comparison}
\begin{tabular}{lcclc}
\toprule
\textbf{Method} &\makecell[c]{\textbf{City} \\ \textbf{Trans.}} & 
\makecell[c]{\textbf{Indicator} \\ \textbf{Gen.}}  &\makecell[c]{\textbf{Data} \\ \textbf{Scale}}  &\makecell[c]{\textbf{Reasoning} \\ \textbf{Ability}}\\
\midrule
SVF&\cmark & \xmark  &$10^4$&\xmark \\
KnowCL      &\cmark & \xmark  &$10^3$&\xmark \\
UrbanCLIP&\cmark & \xmark  &$10^4$&\xmark \\
GeoLLM &\cmark & \xmark  &$10^4$&\xmark \\
UrbanMLLM    &\xmark & \xmark  &$10^5$&\xmark \\
 UI-CoT& \xmark & \xmark & $10^5$&\xmark \\
CityRiSE     &\cmark & \cmark  &$10^3$&\cmark \\
\bottomrule
\end{tabular}
\vspace{-10pt}
\end{table}

In response to these challenges, a number of recent studies have begun to explore learning-based approaches for socio-economic prediction from urban visual data. Early approaches, such as SVF~\cite{fan2023uvi}, uses a two-stage pipeline where visual features are extracted from street view images using a computer vision model and then fed into regression model for socio-economic prediction. More recent efforts explore more integrated approaches based on knowledge graphs and contrastive learning frameworks to better model complex urban patterns~\cite{zhou2023hierarchical, liu2023knowledge, chen2024profiling}.
With the emergence of large vision-language models~\cite{feng2025citygpt,feng2025urbanllava}, these powerful models have also started to be applied to socio-economic indicator prediction tasks~\cite{yan2024urbanclip, hao2025urbanvlp, zhang2025urbanmllm, mi2025econgym,he2025urbanfeel}.
However, as summarized in Table~\ref{tab:method_comparison}, existing approaches face several limitations:
First, while many existing models achieve strong performance on indicators or cities seen during training, their transferability to unseen regions or targets remains limited. Some methods explore cross-city prediction, but with poor performance, and few address the more challenging task of cross-indicator generalization.
Second, existing approaches incur high data costs. Classic methods require large amounts of training data and often rely on external resources such as knowledge graphs. LVLM-based methods depend on supervised fine-tuning (SFT) with tens of thousands of labeled samples, limiting their scalability in low-resource settings.
Third, most models lack interpretability. They output numerical predictions without providing reasoning steps or explanation, making it difficult to analyze or trust the model's decision-making process.

To address these limitations, we propose \textit{CityRiSE}, a reinforcement learning framework built on top of the LVLM for generalist urban socio-economic status prediction. 
CityRiSE first enables generalization to unseen cities and indicators by normalizing diverse socio-economic targets into a unified prediction format and learning transferable reasoning patterns that connect visual input to high-level socio-economic concepts during training.
In addition, we adopt Group Relative Policy Optimization (GRPO) as our training strategy and introduce two auxiliary datasets, the Perceptual Urban Reasoning Data and the General Visual Reasoning Data, that cultivate complementary reasoning skills and help the LVLM develop transferable cognitive abilities. This allows CityRiSE to achieve strong performance using only 5,109 training samples, without relying on large-scale supervised data. 
Finally, to support the emergence of interpretable reasoning, we design a verifiable reward mechanism consisting of a regression reward that enforces numerical correctness and a keyword reward that implicitly encourages the model to generate coherent, goal-directed reasoning chains. The main contributions of this paper are summarized as follows:
\begin{itemize}[leftmargin=1.5em,itemsep=0pt,parsep=0.2em,topsep=0.0em,partopsep=0.0em]
\item We are the first, to our knowledge, to use reinforcement learning with LVLMs to enable emergent visual reasoning for high-level urban socio-economic status prediction.
\item By learning to predict normalized indicator status, CityRiSE acquires transferable reasoning patterns that support strong generalization across cities and indicators, 
with cross-indicator generalization demonstrated for the first time in this domain.
\item By adopting GRPO and constructing auxiliary datasets that target transferable perceptual and reasoning skills, CityRiSE achieves strong performance using only 5k training samples, surpassing even proprietary commercial LVLMs despite the significantly reduced data scale.
\item Through verifiable reward design, CityRiSE induces emergent, coherent, and interpretable reasoning chains for indicator prediction, offering transparency and explanatory insights beyond black-box outputs.
\item Extensive experiments demonstrate that CityRiSE outperforms existing SOTA baselines in 10 urban socio-economic status prediction tasks, particularly for prediction in unseen cities and unseen indicators. Our codes and datasets are open-sourced via \url{ https://github.com/tsinghua-fib-lab/CityRiSE}.
\end{itemize}

%% file: sec2_methods.tex
\section{Methods} \label{sec:methods}
\subsection{Transferable Datasets for Cross-City and Cross-Indicator Generalization} 
To enable reasoning that generalizes across cities and indicators, we construct a dataset suite comprising a primary dataset for socio-economic prediction as well as two auxiliary datasets, which promote transferable cognitive patterns by targeting key perceptual and logical reasoning skills relevant to urban understanding.
\vspace{-5pt}
\subsubsection{Socio-Economic Indicator Data}
We use the CityLens dataset, a multi-modal urban dataset spanning 17 cities worldwide and providing ground-truth annotations for socio-economic indicators across six key urban domains~\cite{liu2025citylens}.
To normalize the heterogeneous value scales across indicators, we follow GeoLLM~\cite{manvi2024geollm} and discretize each indicator into 10 bins, with values ranging from 1 to 10.
Each input consists of one satellite image and ten street view images from a given region, and the model is required to predict a target socio-economic indicator for that region. To evaluate generalization, we consider two complementary settings: (1) spatial transferability, where 10 cities are used for training and evaluation while the remaining 7 are held out for testing, with both splits covering the Global South and Global North 
to ensure geographic diversity and mitigate potential regional bias; and (2) indicator generalization, where 5 indicators are used during training and the remaining 5 are used only for testing.

\label{sec:sec:dataset}
\begin{figure*}[htbp]
    \centering
    \includegraphics[width=0.90\linewidth]{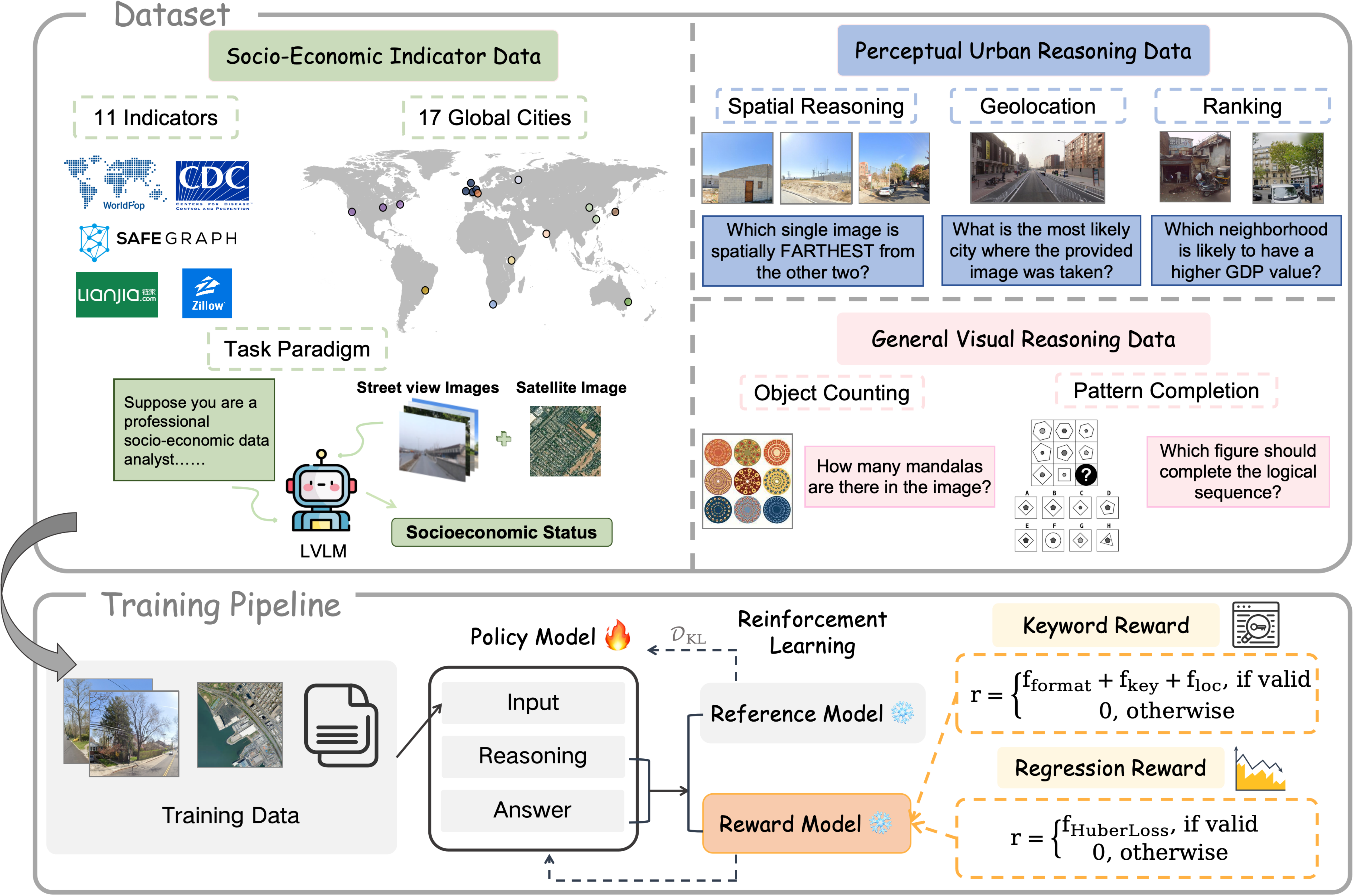}
    \caption{Framework of CityRiSE, which integrates three types of training data with a GRPO training pipeline. The GRPO algorithm incorporates both Keyword Reward and Regression Reward to guide learning.}
    \label{fig:framework}
    \vspace{-10pt}
\end{figure*}
\vspace{-5pt}
\subsubsection{Perceptual Urban Reasoning Data}
To enhance the model's ability to understand and reason about urban environments from street view images, we design three perceptual tasks that capture spatial, geographic, and socio-economic aspects of visual reasoning for training. The images and indicator annotations used to construct this dataset are drawn exclusively from the cities and indicators in the training split, without involving any city-indicator combinations from the test split.

\textit{Spatial Reasoning. }
In this task, each input sample contains three street view images, and the model must determine which image is spatially furthest from the other two. This encourages the model to infer spatial proximity using only visual cues.
We introduce two settings: (1) two images are sampled from the same city, while the third comes from a different city; (2) two images are taken from the same neighborhood or block, while the third comes from a different neighborhood within the same city. These settings require the model to develop a fine-grained perception of urban layouts, building density, architectural styles, and environmental continuity.

\textit{Geolocation Task. }
This task draws on a well-established tradition in urban geospatial research, where inferring a city's identity from visual data has long served as a benchmark for evaluating spatial understanding~\cite{li2024georeasoner,li2025recognition}. Given a single street view image, the model is required to predict its city of origin.  
This setting not only tests the model's ability to extract high-level geographic and architectural signals, like signage, vegetation, or road infrastructure, but also provides valuable priors for socio-economic status prediction.  
By grounding visual inputs in specific geographic contexts, the model can leverage its internally stored world knowledge, such as the average socio-economic profile of the predicted city, to make more informed and context-aware predictions.

\textit{Socio-Economic Ranking. }
Inspired by prior work in image quality assessment that formulates visual comparison as a learning-to-rank problem~\cite{wu2025visualquality, zhang2021uncertainty}, we design a perceptual reasoning task in which the model compares two street view images and determines which region has a higher value for a given socio-economic indicator, relying solely on visual cues.  
This pairwise comparison not only helps the model learn to associate visual patterns with latent urban conditions, but also mirrors the underlying structure of the final prediction objective, which involves distinguishing subtle differences in socio-economic status across regions.

\vspace{-5pt}
\subsubsection{General Visual Reasoning Data}
To support the model's ability in urban socio-economic sensing and abstraction, we incorporate two types of general visual reasoning tasks to build fundamental perceptual skills essential for urban understanding.
The first is an object counting task~\cite{paiss2023countbench}, where the model must determine the number of instances of a given object type within an image. This requires the model to localize, distinguish, and quantify multiple objects under diverse spatial arrangements.
The second is a pattern completion task, inspired by classic geometric reasoning problems~\cite{leng2025mmr1}. In this setting, the model is presented with a visual sequence or matrix with a missing element and is required to infer the underlying rule to select the correct option. These tasks encourage the model to recognize abstract visual patterns, including regularity, symmetry, analogy, and transformation, which are transferable to urban reasoning challenges, such as identifying socio-spatial structures or comparing urban forms across cities.

\vspace{-5pt}
\subsection{Guided Reward Design for Interpretable Socio-Economic Reasoning} \label{sec:sec:reward}
We train the LVLM using the GRPO algorithm, and design a dual-component reward mechanism to guide the optimization process. Table~\ref{tab:data_statis} summarizes the reward types used for different training data. In particular, we introduce two novel reward functions tailored for the socio-economic indicator data. The Keyword Reward guides the LVLM toward more goal-directed perception and generation, thereby facilitating interpretable reasoning processes. The Regression Reward, in turn, promotes outputs that are numerically aligned with the ground-truth values and can be objectively evaluated.
\input{tabs/data_statis}
\vspace{-5pt}
\subsubsection{Keyword Reward}
To guide the model toward producing responses that reflect meaningful urban visual features, we design a keyword-based reward component. This module is integrated into the format reward of GRPO, serving as an auxiliary signal that provides a lightweight perceptual prior during training. Rather than prescribing a fixed reasoning template, it encourages the model to attend to and cover task-relevant visual evidence that is potentially informative for socio-economic prediction.
Concretely, we define a list of six urban-perceptual keywords: ``person'', ``vehicle'', ``greenery'', ``road infrastructure'', ``street furniture'', and ``building''. 
These keywords are derived from prior work on urban visual intelligence~\cite{fan2023uvi}, which extracts 13 visual features using deep learning models for socio-economic prediction. 
We cluster these 13 features into six categories from the perceptual perspective of LVLMs. 
These keywords encompass nearly all visual cues that are visible in street view images and relevant to socio-economic prediction. Guided by these keywords, the model is encouraged to focus on informative visual content while retaining the flexibility to discover and integrate additional cues beyond the predefined keyword set.
In addition, we assign a special reward to the mention of ``location'', which indicates that the model attempts to anchor the visual input to a specific city. This grounding process enables the model to retrieve its internally stored prior knowledge about the city, such as its average socio-economic levels, and integrate it into its reasoning. This step is crucial for models that aim to generalize across diverse urban environments.
\begin{equation}
R_{\text{keyword}}(r) =
\lambda_{\text{base}} \cdot \mathbf{1}_{\text{format}(r)} +
\sum_{k \in \mathcal{K}} \lambda_k \cdot \mathbf{1}_{k \in r} +
\lambda_{\text{loc}} \cdot \mathbf{1}_{\text{loc} \in r}.
\end{equation}
\vspace{-5pt}
\begin{equation}
\mathbf{1}_{\text{condition}} =
\begin{cases}
1, & \text{if the condition holds}, \\
0, & \text{otherwise}.
\end{cases}
\end{equation}

\subsubsection{Regression Reward}
To better reflect the nature of our task as a regression problem, we design a reward function sensitive to the magnitude of prediction errors. In socio-economic indicator prediction, the model outputs a continuous score, and small deviations from the ground truth should not be penalized as harshly as large ones.
For example, if the ground truth is 8, predicting 7 should be rewarded more than predicting 1, a subtlety that binary correctness-based metrics fail to capture.
To address this, we introduce a regression reward based on the Huber loss:
\vspace{-5pt}
\begin{equation}
\text{error} = y_{\text{pred}} - y_{\text{true}}, 
\quad 
|\text{error}| = \left| y_{\text{pred}} - y_{\text{true}} \right|,
\end{equation}
\vspace{-10pt}
\begin{equation}
L_{\text{Huber}}(\text{error}) =
\begin{cases}
\frac{1}{2} \, \text{error}^2, & \text{if } |\text{error}| \leq \delta \\
\delta \, \left( |\text{error}| - \frac{1}{2} \delta \right), & \text{otherwise}
\end{cases},
\end{equation}
\vspace{-15pt}

\begin{equation}
R_{\text{regression}} = \exp \left( -\alpha \, L_{\text{Huber}}(\text{error}) \right).
\end{equation}
This formulation yields a reward that decays smoothly and non-linearly with prediction error. The use of Huber loss ensures stability near the correct value due to its quadratic form, while reducing sensitivity to outliers through its linear behavior for larger errors. The exponential transformation ensures that predictions closer to the true value receive rewards close to 1, while distant predictions are penalized more strongly.
Overall, this reward better aligns with the continuous, fine-grained nature of the prediction task, and provides a more informative learning signal during training. The same reward formulation is used for Object Counting Data, which also involves count-valued outputs.
\vspace{-5pt}
\subsection{GRPO Pipeline for Data-Efficient Learning} \label{sec:sec:training_pipeline}
We train the LVLM using GRPO on the transferable dataset described in Section~\ref{sec:sec:dataset}. As Algorithm~\ref{alg:grpo_algorithm} shows, GRPO operates by generating multiple candidate responses for each data point and calculating the reward for each candidate based on the reward design outlined in Section~\ref{sec:sec:reward}.
To encourage the model to favor higher-reward completions within each group, GRPO computes a group-normalized advantage for each response. This advantage reflects how well each response performs relative to other candidates in the same group. Finally, the model is updated by minimizing a loss function that incorporates both the advantage and the KL divergence, ensuring the model learns to improve its performance on the dataset.
\vspace{-10pt}
\begin{algorithm}[H]
\caption{Training Algorithm for CityRiSE using GRPO}
\label{alg:grpo_algorithm}
\begin{algorithmic}
\STATE {\bfseries Input:} Pretrained policy $\pi_\theta$, reference policy $\pi_{\text{ref}}$, reward model $R$, training dataset $\mathcal{D}$, KL coeff $\beta$, clip coeff $\epsilon$.
\STATE {\bfseries Output:} Optimized policy $\pi_\theta$.
\FOR{each $(x_i, y_i) \in \mathcal{D}$}
\STATE Generate $N$ responses: $\{o_i^{(j)}\}_{j=1}^N \sim \pi_\theta(\cdot \mid x_i)$
\STATE Compute rewards: $r_i^{(j)} = R(o_i^{(j)})$
\STATE Compute advantages:
$\scriptstyle A_i^{(j)} =
\tfrac{r_i^{(j)} - \operatorname{mean}(\{r_i^{(1)}, \ldots, r_i^{(N)}\})}
{\operatorname{std}(\{r_i^{(1)}, \ldots, r_i^{(N)}\})}$
\STATE Compute ratios: $s_i^{(j)} = \displaystyle \frac{\pi_\theta(o_i^{(j)} \mid x_i)}{\pi_{\theta_{\text{old}}}(o_i^{(j)} \mid x_i)}$
\STATE Compute the following objective:
\STATE \parbox[t]{0.92\linewidth}{{\small
\setlength{\jot}{0.7pt}
\(
\begin{aligned}
\mathcal{J}_i
&=\mathbb{E}[\{o_i^{(j)}\}_{j=1}^{N} \sim \pi_{\theta_{\text{old}}}(\cdot \mid x_i)]
\frac{1}{N} \sum_{j=1}^{N}
[\min \big(s_i^{(j)} A_i^{(j)}, \\
&\qquad \operatorname{clip}(s_i^{(j)}, 1-\epsilon, 1+\epsilon) A_i^{(j)}\big)
- \beta \mathcal{D}_{\text{KL}}(\pi_\theta \| \pi_{\text{ref}})]
\end{aligned}
\)
}}
\STATE Update $\pi_\theta$ by maximizing $\mathcal{J}_i$
\ENDFOR
\STATE {\bfseries return} $\pi_\theta$
\end{algorithmic}
\end{algorithm}
\vspace{-5pt}

%% file: tabs/data_statis.tex
\begin{table}[t]
\centering
\caption{Training-task reward schemes and instance statistics.}
\vspace{-10pt}
\setlength{\tabcolsep}{2pt} %
\begin{tabular}{lccc}
\hline
\textbf{Data Type}& \makecell{\textbf{Format} \\ \textbf{Reward}} & \makecell{\textbf{Accuracy} \\ \textbf{Reward}} &\textbf{Instances}\\
\hline
Socio--Economic Indicator & Keyword& Regression &2,828\\
 Spatial Reasoning& Standard& Standard &632\\
 Geolocation Task& Standard& Standard &350\\
Socio-Economic Ranking& Standard& Standard &699\\
Object Counting & Standard& Regression &300\\
Pattern Completion & Standard& Standard &300\\
\hline
\end{tabular}
\label{tab:data_statis}
\vspace{-10pt}
\end{table}

%% file: sec3_experiment.tex
\input{tabs/main_result}
\vspace{-15pt}
\section{Experiments} \label{sec:exp}
To evaluate the effectiveness, generalization ability, and interpretability of CityRiSE, we organize our experiments around the following three research questions:
\begin{itemize}[leftmargin=1.5em,itemsep=0pt,parsep=0.2em,topsep=0.0em,partopsep=0.0em]
\item \textbf{RQ1:} How does CityRiSE perform compared to proprietary LVLMs and SFT-based LVLMs, across in-domain settings, cross-city transferability, and cross-indicator generalization?
\item \textbf{RQ2:} How does the design of the reward mechanism and the composition of training data influence the performance of CityRiSE?
\item \textbf{RQ3:} How well are CityRiSE's specific design choices justified, and what complementary roles do street-view and satellite imagery play at inference time?
\item \textbf{RQ4:} What role does RL-induced reasoning play in shaping CityRiSE's predictive capabilities?
\end{itemize}
\vspace{-5pt}
\subsection{Baselines}
To comprehensively evaluate our approach, we compare it against 8 baselines in three categories: (1) classic methods that do not involve LVLMs, including SVF variants~\citep{fan2023uvi}, UrbanCLIP~\citep{yan2024urbanclip} and UrbanVLP~\citep{hao2025urbanvlp}; (2) general LVLMs, and (3) LVLMs supervised fine-tuned on domain-specific data, consisting of UrbanMLLM~\citep{zhang2025urbanmllm}, trained on the Socio-Economic Indicator Data-Train dataset, and UI-CoT~\citep{zhang2025inequality}, trained on a Chain-of-Thought (CoT) augmented version of the same dataset.

\vspace{-5pt}
\subsection{Overall Performance (RQ1)}
Table~\ref{table:main_result} presents the main results on the Socio-Economic Indicator Data-Test set. The evaluation tasks are broadly categorized into three types based on their relationship to the training distribution: 
\begin{itemize}[leftmargin=1.5em,itemsep=0pt,parsep=0.2em,topsep=0.0em,partopsep=0.0em]
\item Unseen Indicators. These tasks involve indicators that do not appear in the training set. Except for the Life Expectancy task, which involves both unseen cities and unseen indicators, all other tasks in this group share cities with the training set but introduce new indicators.
\item Unseen Cities. The indicators remain the same as in training, but the evaluation is conducted on cities that were not present in the training set.
\item In-domain. Both the cities and indicators in these tasks are consistent with those seen during training. Importantly, the test set is constructed with entirely disjoint samples to prevent any data leakage and ensure a fair evaluation.
\end{itemize}
\vspace{-5pt}
\paragraph{Generalization to Unseen Indicators}
In the unseen indicators setting, CityRiSE achieves the best overall performance, outperforming all baselines across most of the novel socio-economic indicators. This highlights its strong ability to generalize not only to new visual contexts, but also to entirely unseen prediction targets. The model maintains positive $R^2$ scores on all unseen indicators, including challenging ones such as Life Expectancy (0.200) and House Price (0.218), where other models struggle or fail to generalize. 
Notably, traditional deep learning baselines such as SVF, UrbanCLIP and UrbanVLP are absent from this evaluation. This is because such models require a separate predictor for each target variable, making them inherently limited in handling unseen indicators. This limitation underscores a key advantage of LVLMs: their ability to serve as universal predictors for diverse tasks through language-based prompts.
\vspace{-5pt}
\paragraph{Transferability to Unseen Cities}
For unseen cities setting, CityRiSE demonstrates strong spatial transferability. Notably, it achieves the highest performance on Bachelor Ratio with an $R^2$ of 0.286, outperforming SFT-based models, general LVLMs and classic methods. 
Although UrbanVLP performs competitively in the in-domain setting, its performance drops substantially on unseen cities, indicating limited cross-city transferability.
While CityRiSE shows slightly weaker performance on Public Transport, we observe that many test images lack clear transportation-related visual cues, and PT values in these cities are relatively low compared to the training set. These factors may partially limit generalization, though the overall trend remains consistent across tasks.
GPT-4.1-Mini achieves unexpectedly strong results on PT, which we attribute to its conservative predictions that happen to align with the test distribution.
\begin{figure*}[t]
    \centering
    \includegraphics[width=0.98\linewidth]{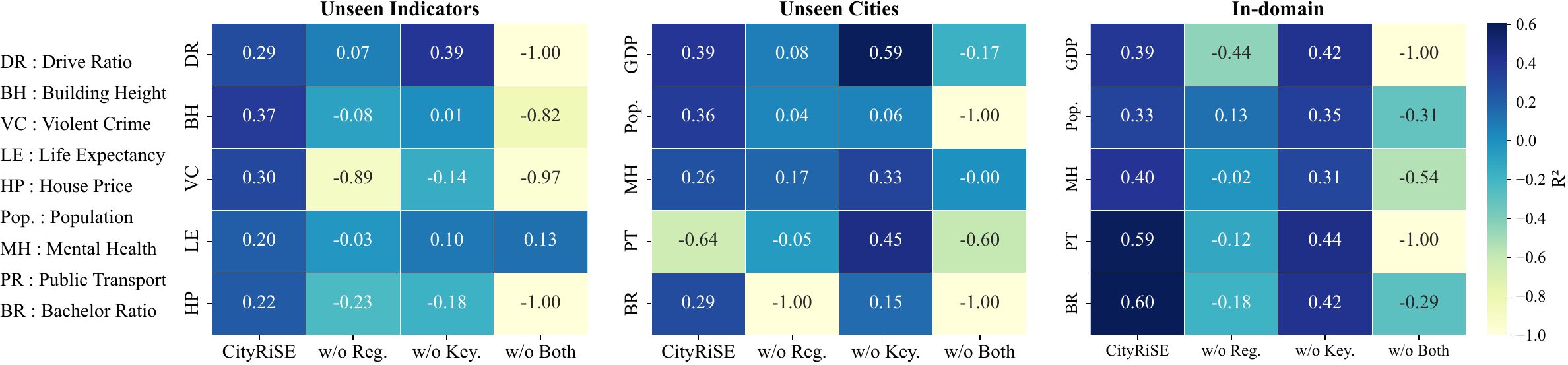}
    \caption{Results of Reward Ablation, where `Reg.' refers to the Regression Reward, `Key.' refers to the Keyword Reward, and `w/o Both' indicates the setting where both rewards are removed.}
    \label{fig:reward_ablation_pic}
    \vspace{-10pt}
\end{figure*}
\vspace{-5pt}
\paragraph{In-domain Analysis}
Under the in-domain setting, CityRiSE significantly outperforms its base model Qwen2.5-VL-7B across all five indicators, with $R^2$ scores improving from 
$-0.330$ to 0.334 on Population and from $-0.013$
 to 0.594 on Public Transport. These results demonstrate the strong predictive capability of CityRiSE. Qwen3-VL-235B A22B Thinking, GPT-4.1-Mini, and Gemini2.5-Flash are all strong general large vision-language models, while their zero-shot performance on socio-economic status prediction remains suboptimal, further highlighting the challenge of this task.
Interestingly, SFT-based models, including UrbanMLLM and UI-CoT, achieve the highest performance on some indicators, particularly GDP and Population. This aligns with the insight proposed in~\citet{chu2025sft} that ``SFT Memorizes, RL Generalizes''. Supervised fine-tuning is highly effective for in-domain memorization, whereas reinforcement learning tends to improve generalization beyond training data.
The comparable performance of SVF-GT and SVF-Rank across in-domain indicators provides empirical support for the validity of our normalization strategy. While not implying full equivalence, the results suggest that normalization preserves predictive accuracy. CityRiSE further surpasses both baselines, demonstrating the benefits of combining reinforcement learning with LVLMs.
\vspace{-5pt}
\subsection{Ablation Study (RQ2)}
\paragraph{Reward Ablation}
Figure~\ref{fig:reward_ablation_pic} shows the results of a reward ablation study comparing the full CityRiSE model with three variants: without Regression Reward, without Keyword Reward, and without both. Removed rewards are replaced by the standard reward. For clarity, $R^2$ values below $-1$ are clipped in the figure.
Removing either reward consistently degrades performance, while removing both leads to severe failure across most indicators. This effect is evident in the in-domain setting, where the $R^2$ score on GDP drops from 0.39 to below $-1.00$, and Public Transport drops from 0.59 to below $-1.00$, indicating insufficient learning signals for effective prediction. The Keyword Reward shows particular importance for specific indicators, such as the drop on Violent Crime from 0.30 to $-0.14$, suggesting its role in enhancing semantic grounding. The Regression Reward, on the other hand, contributes more significantly to numerical precision, as reflected in the drop on Population in unseen cities from 0.36 to 0.04. Overall, the Regression Reward has a greater impact than the Keyword Reward, likely because rewarded keywords are already present in the prompts, reducing the marginal benefit of additional keyword supervision. 
Together, the two rewards offer complementary signals: the Regression Reward supports quantitative alignment, and the Keyword Reward strengthens interpretability and semantic relevance. 

\begin{figure}[h]
    \centering
    \includegraphics[width=0.99\linewidth]{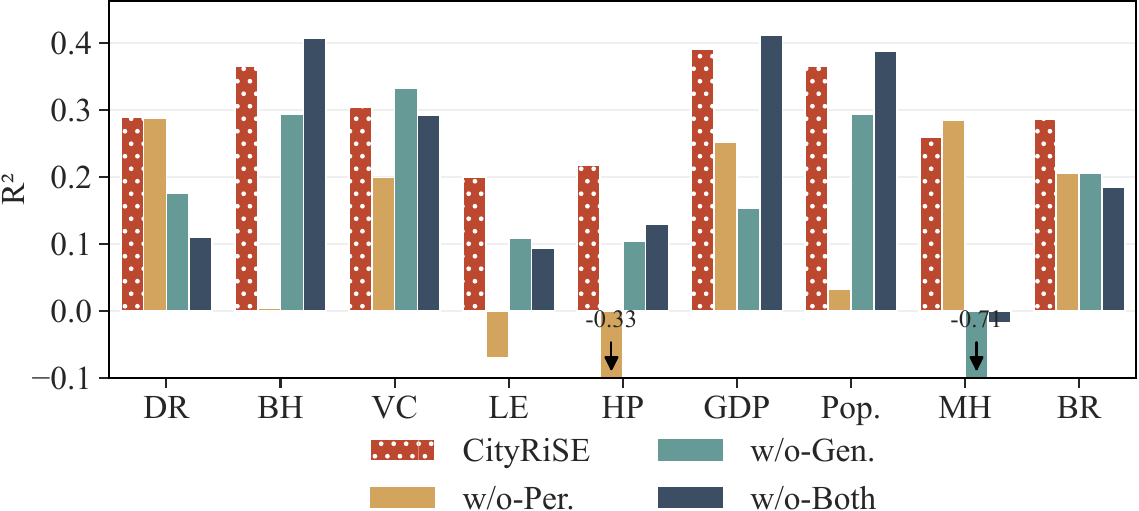}
    \vspace{-5pt}
    \caption{Results of Data Ablation, where `Per.' refers to the Perceptual Urban Reasoning Data, `Gen.' refers to the General Visual Reasoning Data.}
    \label{fig:data_ablation_pic}
    \vspace{-10pt}
\end{figure}

\vspace{-5pt}
\paragraph{Data Ablation}
We incorporate the Perceptual Urban Reasoning Data and General Visual Reasoning Data to enhance the model's reasoning capabilities relevant to socio-economic prediction, particularly its generalization to out-of-domain settings. Figure~\ref{fig:data_ablation_pic} shows the performance of models trained with different data ablations on the unseen indicators and unseen cities subsets. For clarity, tasks where all configurations yield negative $R^2$ scores are excluded, as they offer limited analytical value. 
As shown in Figure~\ref{fig:data_ablation_pic}, removing either auxiliary dataset consistently degrades performance across most indicators, confirming the effectiveness of both Perceptual and General Visual Reasoning data. The General Visual Reasoning Data has broad influence, with its removal causing noticeable drops on nearly all indicators, especially Mental Health, where the $R^2$ score falls from 0.26 to $-0.71$. In contrast, the Perceptual Urban Reasoning Data plays a more targeted role; its removal notably impacts Life Expectancy and House Price, suggesting that spatial comparison and urban identity reasoning support accurate estimation for region-grounded indicators. Removing both datasets leads to severe degradation across nearly all tasks, including complete collapse on Mental Health, highlighting their complementary roles in enabling robust visual reasoning and generalization.

\begin{figure}[ht]
  \centering
  \begin{subfigure}[t]{0.45\columnwidth} 
    \centering
    \includegraphics[width=\textwidth,height=3.1cm]{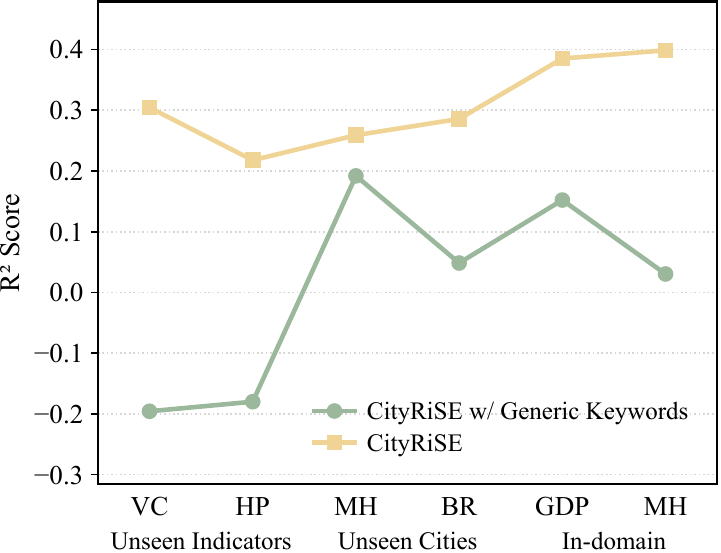} %
    \vspace{-15pt}
    \caption{}
    \vspace{-5pt}
    \label{fig:generic_keyword}
  \end{subfigure}%
  \hfill
  \begin{subfigure}[t]{0.48\columnwidth} 
    \centering
    \includegraphics[width=\textwidth,height=3.1cm]{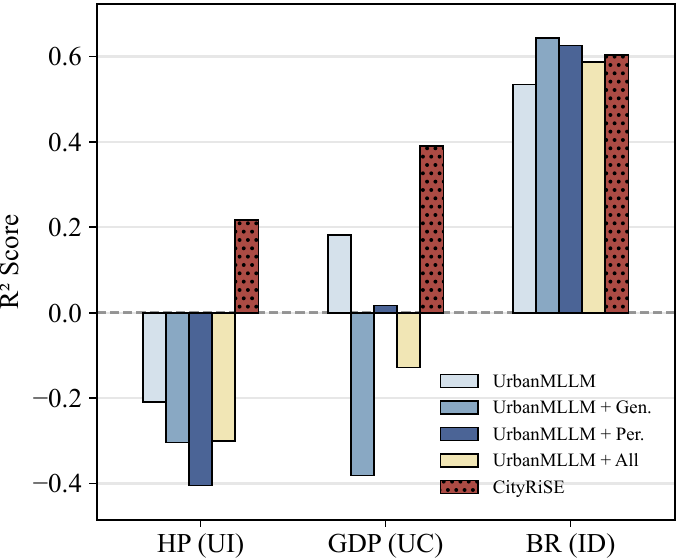} %
    \vspace{-15pt}
    \caption{}
    \vspace{-5pt}
    \label{fig:sft_data_composition}
  \end{subfigure}
  \vspace{-5pt}
  \caption{(a) Comparison between CityRiSE and CityRiSE w/ Generic Keywords. (b) Performance of SFT-LVLMs trained with different data compositions.}
  \vspace{-10pt}
\end{figure}
\vspace{-5pt}
\subsection{Training Design Validation and Input Modality Analysis (RQ3)}
\paragraph{Task-Aligned vs. Generic Keyword Reward}
To further validate the design of the keyword reward, we compare our task-aligned visual keyword set with an alternative keyword design based on generic reasoning-oriented tokens. Specifically, we replace the six visual keywords in CityRiSE with generic thinking tokens, motivated by prior work~\cite{qian2025thinkingtokens} suggesting that expressions such as ``Hmm'', ``Wait'', and ``Therefore'' are crucial for the reasoning performance of large reasoning models. We denote the resulting variant as CityRiSE w/ Generic Keywords. 
As shown in Figure~\ref{fig:generic_keyword}, CityRiSE consistently outperforms this variant on all 6 tasks across three settings. These results indicate that the effectiveness of the keyword reward does not simply come from encouraging deliberative language, but from rewarding task-aligned visual concepts that are directly relevant to urban socio-economic reasoning. This further supports the validity of our reward design.

\vspace{-5pt}
\paragraph{SFT with Different Data Compositions}
We further study how different training data compositions affect SFT performance. Specifically, we conduct supervised fine-tuning on Qwen2.5-VL-7B using the socio-economic indicator data alone and variants further augmented with the auxiliary datasets. Figure~\ref{fig:sft_data_composition} presents the results on 3 tasks spanning the in-domain, unseen cities, and unseen indicators settings. As auxiliary datasets are added, UrbanMLLM shows modest gains in the in-domain setting, whereas performance on most out-of-domain tasks declines, with only a few individual tasks improving. This pattern further confirms the strong in-domain nature of SFT, and suggests that auxiliary data alone is insufficient to yield the transferable gains achieved by CityRiSE.
\vspace{-5pt}
\paragraph{Test-Time Modality Contribution}
\input{tabs/modality_effect}
To examine how CityRiSE utilizes different visual sources, we remove either satellite (SAT) or street-view (STV) imagery at inference time without retraining the model. As shown in Table~\ref{tab:modality_analysis}, removing SAT leads to consistent but moderate degradation across all tasks, with the largest drops observed on MH ($0.259 \rightarrow 0.152$) and HP ($0.218 \rightarrow 0.154$). In contrast, removing STV causes substantially larger declines, producing negative $R^2$ scores on HP, MH, and PT. These results suggest that, under our input configuration, street-view imagery provides the primary fine-grained visual evidence for socio-economic estimation, while satellite imagery supplies complementary regional-scale context. The consistently superior results of the complete model further demonstrate the benefit of combining both modalities.
\vspace{-5pt}
\subsection{Reasoning Analysis and Comparison (RQ4)}
\label{sec:reasoning}
\begin{figure}[ht]
  \centering
  \begin{subfigure}[t]{0.45\columnwidth} 
    \centering
    \includegraphics[width=\textwidth,height=3.1cm]{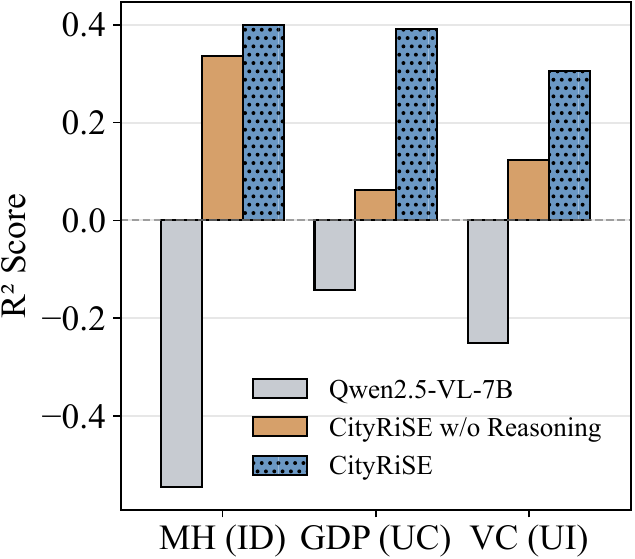} %
    \vspace{-15pt}
    \caption{}
    \vspace{-5pt}
    \label{fig:reasoning_ablation}
  \end{subfigure}%
  \hfill
  \begin{subfigure}[t]{0.48\columnwidth} 
    \centering
    \includegraphics[width=\textwidth,height=3.1cm]{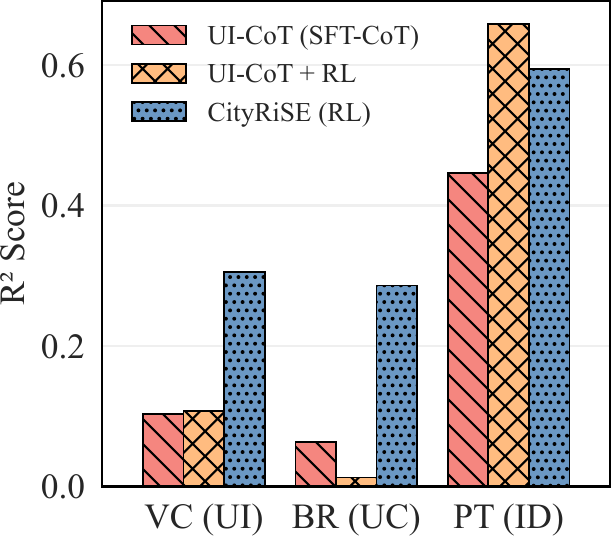} %
    \vspace{-15pt}
    \caption{}
    \vspace{-5pt}
    \label{fig:reasoning_induce}
  \end{subfigure}
  \caption{(a) Comparison of CityRiSE, its no-reasoning RL variant, and the base model across representative tasks. (b) Performance of models trained with different methods on three types of tasks.}
  \vspace{-15pt}
\end{figure}

\paragraph{Reasoning Impact}
To isolate the contribution of reasoning, we construct a no-reasoning variant that keeps the GRPO training procedure, regression reward, and auxiliary data unchanged, but directly generates the final prediction without intermediate reasoning. As shown in Figure~\ref{fig:reasoning_ablation}, this variant outperforms the base model but remains consistently below CityRiSE across all three evaluation settings. These results distinguish the benefit of RL training from that of reasoning and demonstrate that the reasoning process itself contributes to more accurate socio-economic prediction.
\paragraph{Reasoning Integration}
Figure~\ref{fig:reasoning_induce} compares models trained with three different strategies on 3 representative tasks spanning three evaluation settings. 
The three models differ in how reasoning is introduced during training. UI-CoT is fine-tuned on manually constructed CoT annotations; UI-CoT + RL adds reinforcement learning on top of it. In contrast, CityRiSE is trained purely via RL, without access to human-crafted reasoning chains, relying instead on reward design to induce emergent reasoning behavior.
UI-CoT tends to generate rigid, template-like reasoning that works in-domain but generalizes poorly.
RL fine-tuning improves its in-domain results but brings limited out-of-domain gains, likely due to the fixed reasoning style inherited from supervised training, which remains largely unchanged.
In contrast, CityRiSE develops reasoning organically during RL training, guided by verifiable rewards. This results in more flexible and transferable inference patterns, with strong performance on both unseen cities and indicators. These findings support the hypothesis that emergent, reward-induced reasoning is more adaptable to novel contexts than explicitly scripted approaches.
We further apply RL to UrbanMLLM, a SFT-only model. Despite prompting it to ``Please reason step-by-step and write your reasoning inside the <think> </think> tags'' during RL, the model consistently outputs only the final answer. This suggests that if reasoning is not introduced during the early SFT stage, the model may become too rigid for RL to induce reasoning capability later.

\vspace{-5pt}
\begin{figure*}[htbp]
    \centering
    \includegraphics[width=0.96\linewidth]{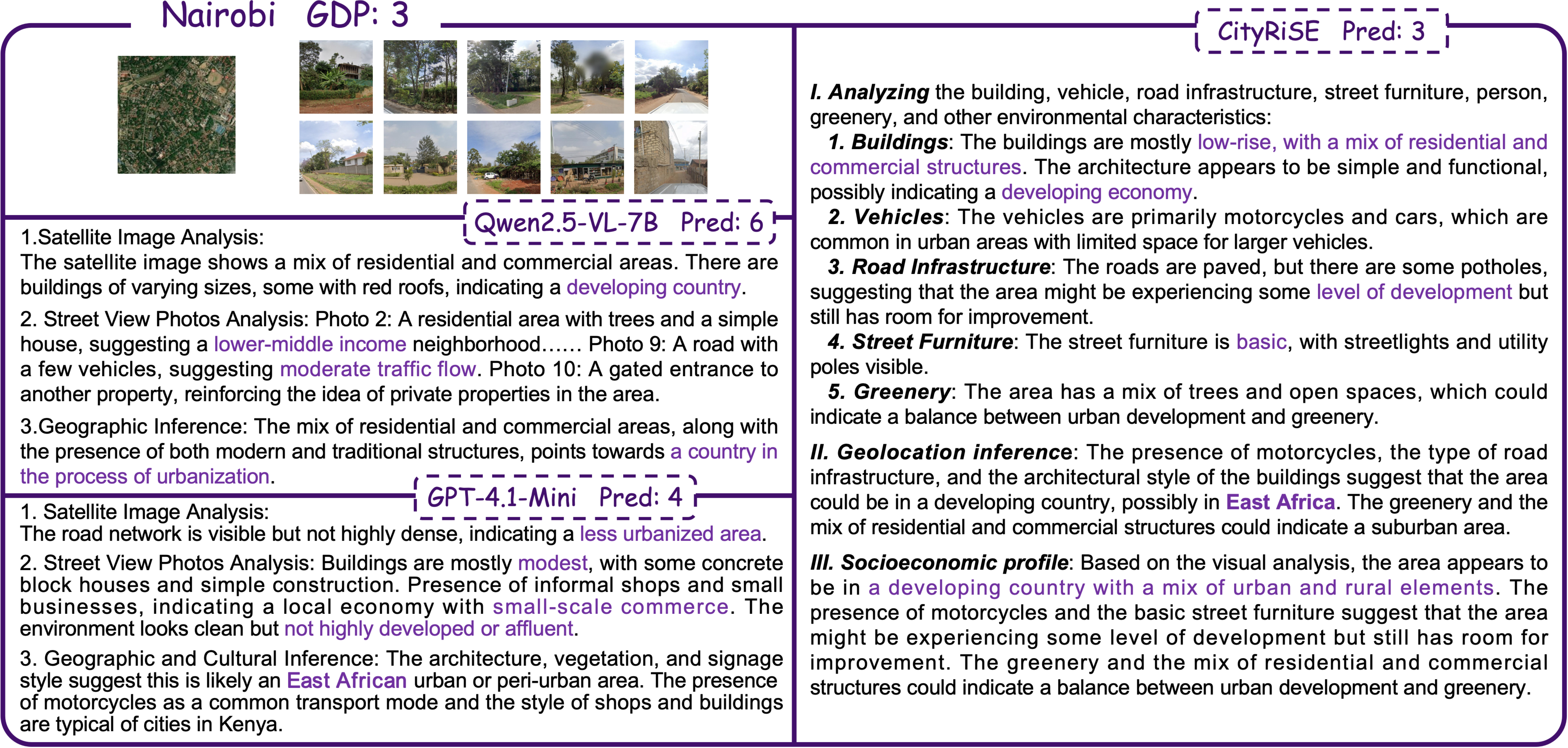}
    \caption{Qualitative comparison of reasoning patterns produced by different models. CityRiSE exhibits more interpretable and goal-directed reasoning behavior, while baseline models often generate less informative responses.}
    \label{fig:reasoning_case}
    \vspace{-5pt}
\end{figure*}

\paragraph{Reasoning Case}
Figure~\ref{fig:reasoning_case} presents a qualitative comparison of partial reasoning chains generated by three models for the same socio-economic prediction task. Although CityRiSE is trained via reinforcement learning without access to answer labels, it develops a clear and coherent reasoning structure. This emerges from the combined effect of the Keyword Reward and prompt design, which offer implicit and explicit guidance toward goal-directed reasoning. 
Compared to its base model (Qwen2.5-VL-7B), CityRiSE demonstrates a more holistic and higher-level perception across multiple images, rather than producing low-information summaries of each image in isolation. 
Moreover, CityRiSE is able to more accurately perceive and 
reason about the street view images. It successfully infers the likely location as ``East Africa'', aligning with the output of GPT-4.1-Mini, a much larger proprietary model. This geographic inference provides valuable prior knowledge for socio-economic status prediction, as regional context can help calibrate the model's expectations for certain indicators.
These results demonstrate not only robust visual grounding, but also a natural and uninterrupted reasoning process, emerging organically through RL training.

%% file: tabs/main_result.tex
\begin{table*}[t]
\centering
\setlength{\tabcolsep}{3pt} %
\footnotesize %
\renewcommand\arraystretch{1.1} %
\caption{Main results on Socio-Economic Indicator Data-Test. The evaluation metric is the coefficient of determination ($R^2$). In each row, bold indicates the best result, and underline denotes the second-best.}
\vspace{-5pt}
\label{table:main_result}
\resizebox{\textwidth}{!}{
\begin{tabular}{llllllllllllccccc} 
\toprule
\textbf{Domain} &{\cellcolor{lightred}\textbf{Overall}} & \multicolumn{5}{c}{\cellcolor{lightyellow}\textbf{Unseen Indicators}} & \multicolumn{5}{c}{\cellcolor{lightblue}\textbf{Unseen Cities}}& \multicolumn{5}{c}{\cellcolor{lightgray}\textbf{In-domain}}\\
\textbf{Tasks}  & {\cellcolor{lightred}}& \cellcolor{lightyellow}\textbf{DR}& \cellcolor{lightyellow} \textbf{BH}& \cellcolor{lightyellow}\textbf{VC}& \cellcolor{lightyellow}\textbf{LE}&\cellcolor{lightyellow}\textbf{HP}  & \cellcolor{lightblue} \textbf{GDP}& \cellcolor{lightblue}\textbf{Pop.}& \cellcolor{lightblue} \textbf{MH}& \cellcolor{lightblue}\textbf{PT}&\cellcolor{lightblue} \textbf{BR} & \cellcolor{lightgray}\textbf{GDP}& \cellcolor{lightgray}\textbf{Pop.}& \cellcolor{lightgray}\textbf{MH}& \cellcolor{lightgray} \textbf{PT}&\cellcolor{lightgray}\textbf{BR} \\ 
\cmidrule(lr){1-17}
\textbf{Random} &-0.929& -0.916 & -0.734 & -1.106 & -0.941 &-0.951  & -2.742& -0.874 & -2.902 & -4.694&-2.085 & -1.293 & -1.220 & -0.798 &  -0.562 &-1.191 \\
\textbf{SVF-GT} &/ & /& /& /& /&/ & -1.046& 0.091 & -1.964 & -1.358&-0.517 & 0.329 & 0.268 & 0.087 &  0.417 &0.210 \\
\textbf{SVF-Rank} &/ & / & / & /& /&/  & -0.560& 0.172 & -2.154 & -1.581&-0.720 & 0.366 & 0.267 & 0.162&  0.358 &0.287 \\
\textbf{UrbanCLIP} &/ & /& /& /& /&/ & -0.595& 0.161& -1.537& -3.137&-0.622& 0.176& 0.015& -0.015&  -0.025&0.008\\
\textbf{UrbanVLP} &/ & / & / & /& /&/  & -0.279& 0.146& -0.174& -0.452&-0.265& \uline{0.609}& \uline{0.500}& \uline{0.260}&  0.568&0.510\\
\midrule
\textbf{Qwen2.5-VL-7B} &-0.116& -0.130 & 0.178 & -0.251 & \uline{0.083} &-0.026  & -0.143& -0.209 & -2.172 & -0.118&-0.918 & -0.280 & -0.330 & -0.545 &  -0.013 &-0.094 \\
\textbf{Qwen3-VL-235B} &-0.572& -1.534 & 0.290 & -1.150 & -0.874 &0.056  & 0.114& -0.405 & -5.924 & 0.004&-0.169 & -0.752 & -0.944 & -1.288 &  -0.621 &0.173 \\
\textbf{GPT-4.1-Mini} &-0.233& \uline{0.252} & 0.285 & -0.908 & -0.023 &\uline{0.111}  & \textbf{0.444}& -0.183 & -4.943 & \textbf{0.413}&-1.386 & -0.323& -0.222 & -0.989 &  0.342 &0.239 \\
\textbf{Gemini2.5-Flash} &-0.718& -0.903 & \uline{0.395} & -1.435 & -0.525 &-0.211  & -0.260& -0.529 & -6.709 & \uline{0.098}&-1.801 & -0.825 & -0.918 & -1.078 &  0.529 &-0.220 \\
\midrule
\textbf{UrbanMLLM (SFT)} &-0.007& -2.249 & 0.314 & -0.087 & -1.400 &-0.209  & 0.182& \uline{0.274} & \textbf{0.329} & 0.067&-0.946 & \textbf{0.691}& \textbf{0.574} & -0.005 & \textbf{0.607} &0.534 \\
\textbf{UI-CoT (SFT-CoT)} &\uline{0.176}& -0.939 & \textbf{0.398} & \uline{0.103} & -0.493 &-0.105  & -0.419& -0.055 & \uline{0.264} & -0.073&\uline{0.063} &0.546& 0.499& 0.053 &  0.446 &\textbf{0.616} \\
\midrule
 \textbf{CityRiSE (only RL)} &\textbf{0.421}& \textbf{0.289} & 0.366 & \textbf{0.305} & \textbf{0.200} &\textbf{0.218}  & \uline{0.391}& \textbf{0.365} & 0.259 & -0.635&\textbf{0.286} &0.385 &0.334 &\textbf{0.399} &\uline{0.594} & \uline{0.603} \\
\bottomrule
\end{tabular}}
\vspace{-10pt}
\end{table*}

%% file: tabs/modality_effect.tex
\begin{table}[t]
    \centering
    \caption{Performance of CityRiSE under different test-time input-modality settings, compared with its base model.}
    \vspace{-10pt}
    \label{tab:modality_analysis}
    \setlength{\tabcolsep}{3.2pt}
    \renewcommand{\arraystretch}{1.03}
    \scriptsize
    \resizebox{\columnwidth}{!}{
    \begin{tabular}{@{}lcccccc@{}}
        \toprule
        \multirow{2}{*}{\textbf{Task}}
        & \multicolumn{2}{c}{\textbf{UI}}
        & \multicolumn{2}{c}{\textbf{UC}}
        & \multicolumn{2}{c}{\textbf{ID}} \\[-2pt]
        & \textbf{DR}
        & \textbf{HP}
        & \textbf{Pop.}
        & \textbf{MH}
        & \textbf{PT}
        & \textbf{BR} \\
        \midrule

        \textbf{CityRiSE}
        & $\mathbf{0.289}$ & $\mathbf{0.218}$
        & $\mathbf{0.365}$ & $\mathbf{0.259}$
        & $\mathbf{0.594}$ & $\mathbf{0.603}$ \\

        \textbf{w/o SAT}
        & $\underline{0.270}$ & $\underline{0.154}$
        & $\underline{0.328}$ & $\underline{0.152}$
        & $\underline{0.579}$ & $\underline{0.583}$ \\

        \textbf{w/o STV}
        & $0.011$ & $-0.277$
        & $0.147$ & $-0.658$
        & $-0.202$ & $0.157$ \\
        \bottomrule
    \end{tabular}
    }
    \vspace{-15pt}
\end{table}

%% file: sec4_related_work.tex
\vspace{-5pt}
\section{Related Work} \label{sec:related}
\paragraph{Socio-Economic Prediction from Urban Imagery}
In general, Urban socio-economic sensing~\cite{kang2020review} aims to predict high-level indicators based on visual data from urban environments, particularly satellite and street view images.
The core challenge is effectively extracting useful multi-source information from these multi-modal data for accurate assessment. To address this challenge, researchers have developed various modeling approaches that leverage the diverse, multi-modal urban data, ranging from classic regression methods~\cite{fan2023uvi} to modern deep learning~\cite{liu2023knowledge,zhou2023hierarchical} and representation learning~\cite{chen2024profiling, yong2024musecl} techniques. 
With the advent of LLMs, a promising new direction involves leveraging their embedded commonsense and world knowledge to enrich urban visual data with contextual information~\cite{li2024can,feng2025citybench,liu2025citylens}. 
The typical paradigms for this integration fall into two categories: using LLMs as a supplementary tool for extra information to assist the aforementioned classical methods \cite{yan2024urbanclip,hao2025urbanvlp}, and directly employing LLMs as the predictor by leveraging their rich internal knowledge and modeling capabilities. The second approach, which more tightly integrates the LLM's inherent knowledge with indicator prediction, has garnered more attention~ \cite{zhou2024synergizing,manvi2024large,han2024geosee}.
GeoLLM \cite{manvi2024geollm} links the LLM's internal knowledge to prediction indicators via address information, and UrbanMLLM \cite{zhang2025urbanmllm} completes multi-type and multi-task indicator prediction directly through large-scale data pre-training and supervised learning.

\vspace{-5pt}
\paragraph{Generalist Reasoning in LVLMs}
Prior work primarily guides LVLMs through fixed prompt design for visual reasoning \cite{guo2024regiongpt,cheng2024spatialrgpt,wu2024v,xu2026visulogic}. More recently, autonomous visual reasoning in LVLMs based purely on Reinforcement Learning has become a popular topic. The core goal is to train LVLMs via RL, to independently learn to analyze and reason, thereby boosting performance and crucially generalization ability in typical visual tasks \cite{wang2025multimodal,su2025thinking,meng2025mm,wei2026open}. Current work in this space has mainly concentrated on the understanding and reasoning of low-level, intuitive visual elements, such as mathematical geometry problems and natural image comprehension. By utilizing datasets annotated with explicit reasoning processes \cite{xu2024llava,shen2025fine,huang2026vision,deng2025openvlthinker} and carefully designed reward functions \cite{shen2025vlm,xiao2025advancing,xia2025visionary}, similar to the approach of DeepSeek-R1~\cite{guo2025deepseek}, researchers guide LVLMs toward more general visual reasoning, leading to higher performance and better generalization. Distinct from these efforts focusing on intuitive, low-level visual elements, our work is the first to apply the RL paradigm to implicit, high-level visual reasoning. We achieve the association and mining of intuitive elements with hidden knowledge, thus enabling the generalized prediction of unseen socio-economic indicators for the first time while simultaneously enhancing the generalized prediction capability for unseen cities.